\documentclass{article}
\usepackage{spconf,amsmath,graphicx}
\usepackage{color}

\usepackage{multirow}
\usepackage{array}
\usepackage{booktabs}

\title{Picking the Underused Heads:\\a Network Pruning Perspective of Attention Head Selection\\for Fusing Dialogue Coreference Information}

\name{Zhengyuan Liu, Nancy F. Chen\vspace{-0.3cm}}
\address{Institute for Infocomm Research, A*STAR, Singapore\vspace{-0.2cm}}

\begin{document}

\maketitle

\begin{abstract}
The Transformer-based models with the multi-head self-attention mechanism are widely used in natural language processing, and provide state-of-the-art results. While the pre-trained language backbones are shown to implicitly capture certain linguistic knowledge, explicitly incorporating structure-aware features can bring about further improvement on the downstream tasks. However, such enhancement often requires additional neural components and increases training parameter size.
In this work, we investigate the attention head selection and manipulation strategy for feature injection from a network pruning perspective, and conduct a case study on dialogue summarization. We first rank attention heads in a Transformer-based summarizer with layer-wise importance. We then select the underused heads through extensive analysis, and inject structure-aware features by manipulating the selected heads. Experimental results show that the importance-based head selection is effective for feature injection, and dialogue summarization can be improved by incorporating coreference information via head manipulation.
\end{abstract}


\section{Introduction}
\label{sec:introduction}
Recently, the Transformer-based models have shown state-of-the-art performance across a variety of Natural Language Processing (NLP) tasks, including, but not limited to, machine translation and reading comprehension \cite{dasigi-etal-2019-quoref}.
One key component of the Transformer architecture \cite{vaswani-2017-Transformer} is the layer stacking of multi-head attention that allows the model to capture both local and global information to build feature-rich contextualized representations. In large-scale pre-trained language backbones, it is shown that attention heads at different layers play different roles, and potentially capture grammatical features such as part-of-speech \cite{tenney2018what} and structural information like syntactic dependency \cite{hewitt-manning-2019-structural}.
However, without directly training on corpora that provide explicit and specific linguistic annotation such as coreference and discourse information, model performance remains subpar for language generation tasks that require high-level semantic reasoning \cite{dasigi-etal-2019-quoref}. Thus, incorporating such features in a more explicit way raises emerging research interest \cite{xu-etal-2020-discourse,liu-etal-2021-coreference}, including adding attention constrain \cite{wang-etal-2019-treeTransformer} and adopting separate graph neural components \cite{xu-etal-2020-discourse}.

When fine-tuned on downstream tasks, previous studies show that Transformer-based models are over-parameterized, and can be compressed via structured pruning or knowledge distillation. For instance, previous work showed that a few attention heads do the ``heavy lifting'' whereas others contribute very little or nothing at all \cite{voita-etal-2019-Heavy}. In practice, in a well-trained multi-layer Transformer, a large percentage of attention heads can be masked at the inference stage without significantly affecting performance, and some layers can even be reduced to only one head \cite{michel2019sixteen}. Inspired by this observation, we rethink the strategy of incorporating structure-aware information in a network pruning perspective: redundant attention heads can be replaced with featured weights, and it is much more computationally efficient than introducing additional neural components. 
In this paper, we conduct a case study on abstractive dialogue summarization, where high-level semantic features are necessary for achieving optimal performance. We investigate the following two research questions:
\begin{itemize}
\item Are some attention heads less important or redundant in a well-trained dialogue summarizer?
\item Can we manipulate the underused heads with coreference information to improve the summarization model?
\end{itemize}

\noindent Experiments are conducted on a benchmark dialogue summarization corpus SAMSum \cite{gliwa-etal-2019-samsum}. We first take a quantitative observation on the importance of attention heads by scoring and ranking them with a gradient-based algorithm, and conducting structured head pruning at the fine-tuning and inference stage. We empirically find that masking a set of lowest-ranking heads does not affect the model performance on the downstream task, regardless of different training settings.
We then evaluate two head manipulation methods to incorporate the coreference information into the neural dialogue summarizer, and experimental results show that the model can obtain improved performance, and the manipulated heads are effectively utilized with higher importance.

\section{Transformer-based Models}
\label{sec:transformer}
The Transformer \cite{vaswani-2017-Transformer} utilizes self-attention instead of recurrent or convolutional neural components. It can be in the form of encoder-only and encoder-decoder architectures, and Transformer-based sequence-to-sequence models are popular in various language generation tasks such as machine translation and summarization \cite{lewis-etal-2020-bart}.
The encoder consists of stacked Transformer layers, and in each of them, there are two sub-components: a multi-head self-attention layer and a position-wise feed-forward layer. Between these two sub-components, residual connection and layer normalization are added. The $u$-th encoding layer is formulated as:
\begin{equation}
\label{eq:attn-head}
\widetilde{h}^u=\mathrm{LayerNorm}(h^{u-1}+\mathrm{MultiHeadAttn}(h^{u-1}))
\end{equation}
\vspace{-0.5cm}
\begin{equation}
\label{eq:attn-residual}
h^u=\mathrm{LayerNorm}(\widetilde{h}^u+ \mathrm{FFN}(\widetilde{h}^u))
\end{equation}
where $h^u$ is the input to $u$-th layer. $\mathrm{MultiHeadAttn}$, $\mathrm{FNN}$, and $\mathrm{LayerNorm}$ are multi-head attention, feed-forward, and layer normalization, respectively.
The decoder consists of stacked Transformer layers as well. In addition to the two sub-components in encoder, the decoder performs another multi-head attention over the previous decoding hidden states, and over all encoded representations (i.e., cross-attention). Generally, the decoder generates tokens in an auto-regressive manner from left to right.

One sophisticated design of the Transformer for a strong representation learning capability is the multi-head attention mechanism. More specifically, instead of performing a single attention calculation on the input tuple (\emph{i.e.,} key, value, and query) in a $d$-dimension, multi-head attention projects them into $N_h$ different sub-spaces (each sub-space is expected to capture different features \cite{vaswani-2017-Transformer,hewitt-manning-2019-structural}). After calculating attention for every head, where produces a $d/N_h$-dimensional output, we aggregate and project the vectors, and obtain the final contextualized representation. The multi-head attention of one layer is formulated as:
\begin{equation}
\small
    \mathrm{Attention}(Q,K,V) = \mathrm{Softmax}(\frac{QK^T}{\sqrt{d/N_h}})V
\end{equation}
\vspace{-0.15cm}
\begin{equation}
\small
    head_i = \mathrm{Attention}(QW_i^Q,KW_i^K,VW_i^V)
\end{equation}
\vspace{-0.15cm}
\begin{equation}
\small
    \mathrm{MultiHeadAttn}(Q,K,V) = \mathrm{Concat}(head_1,...,head_{N_h})
\end{equation}

\section{Dialogue Summarization}
\label{sec:dialogue_summarization}
\begin{figure}[t!]
\begin{center}
\includegraphics[width=0.88\linewidth]{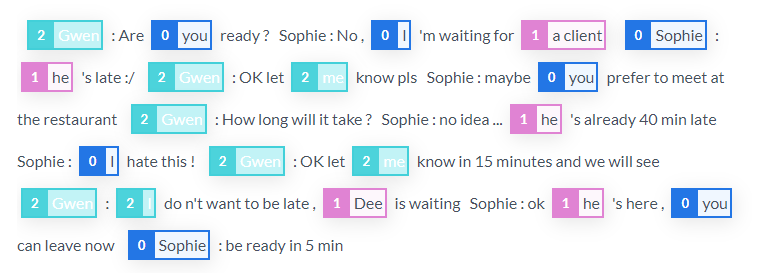}
\end{center}
\vspace{-0.5cm}
\caption{A dialogue example processed by coreference resolution. Colored spans are items in coreference clusters with ID numbers. Each cluster is in one color for better readability.}
\label{fig:dialogue_example}
\vspace{-0.2cm}
\end{figure}

Abstractive dialogue summarization has raised much research interest in recent years \cite{liu2019topic,feng2021survey}.
Unlike documents, conversations are interactions among multiple speakers, they are less structured and are interspersed with more informal linguistic usage \cite{SACKS19787,jurafsky2008speech}, making dialogue summarization more challenging.
In common human-to-human conversations, the useful information (which usually focuses on some dialogue topics) is exchanged back and forth across multiple speakers (\emph{i.e.}, interlocutors) and dialogue turns. Aside from speakers referring to themselves and each other, they also mention third-party persons, concepts, and objects, resulting in ubiquitous coreferential expressions \cite{liu-etal-2021-coreference}. Moreover, the implicit referring such as anaphora or cataphora makes coreference chains more elusive to track.
For instance, as the dialogue shown in Figure \ref{fig:dialogue_example}, two speakers exchange information among interactive turns, where the pronoun \textit{``he''} is used multiple times, referring to the word \textit{``client''}. Without sufficient modeling of the referring information, a summarizer cannot link mentions with their antecedents, and produces outputs with factual inconsistency \cite{tang-etal-2022-confit}. Therefore, enhancing the model with coreference information is beneficial for dialogue summarization to more appropriately context comprehension, and precisely track the interactive flow throughout a conversation. In this work, we conduct a case study on abstractive dialogue summarization, and assess our proposed method of improving context understanding by incorporating conference features.

\begin{table}
\centering
\resizebox{0.93\linewidth}{!}{ \begin{tabular}{p{3.1cm}p{5.3cm}p{2cm}} \toprule \multirow{3}{*}{\begin{tabular}[c]{@{}l@{}}Training Set\\(14,732 Samples)\end{tabular}} & Mean/Std. of Dialogue Turns & 11.7 (6.45) \\ & Mean/Std. of Dialogue Length & 124.5 (94.2) \\ & Mean/Std. of Summary Length & 23.44 (12.72) \\ \midrule \multirow{3}{*}{\begin{tabular}[c]{@{}l@{}}Validation Set\\(818 Samples)\end{tabular}} & Mean/Std. of Dialogue Turns & 10.83 (6.37) \\ & Mean/Std. of Dialogue Length & 121.6 (94.6) \\ & Mean/Std. of Summary Length & 23.42 (12.71) \\ \midrule \multirow{3}{*}{\begin{tabular}[c]{@{}l@{}}Test Set\\(819 Samples)\end{tabular}} & Mean/Std. of Dialogue Turns & 11.25 (6.35) \\ & Mean/Std. of Dialogue Length & 126.7 (95.7) \\ & Mean/Std. of Summary Length & 23.12 (12.20) \\ \bottomrule \end{tabular}}
\caption{\label{table-data-detail}Data statistics of the SAMSum corpus.}
\label{table:corpus}
\vspace{-0.3cm}
\end{table}

\section{Experimental Setting}
\noindent \textbf{Corpus}
In our setting, we conduct experiments on the SAMSum \cite{gliwa-etal-2019-samsum}, a benchmark dialogue summarization dataset consisting of 16,369 daily conversations with human-written summaries. Dataset statistics are listed in Table \ref{table:corpus}.

\noindent \textbf{Evaluation Metrics}
We quantitatively evaluated model outputs with the standard metric ROUGE \cite{lin-och-2004-automatic}, and reported ROUGE-1, ROUGE-2, and ROUGE-L. All reported results use the same evaluation criteria following previous works \cite{gliwa-etal-2019-samsum,feng2021survey} for the benchmarked comparison.

\noindent \textbf{Model Configuration}
The baselines and proposed models were implemented in PyTorch and Hugging Face Transformers. AdamW optimizer was used, and the initial learning rate was set at 1e-5. Beam search size was 5. We trained each model for 15 epochs and selected the best checkpoints on the validation set with ROUGE-2 score. All experiments were run on a single Tesla A100 GPU with 40GB memory.

\section{Head Importance for Specific Downstream Tasks}
To assess the importance of attention heads in a Transformer-based model, we rank the heads from a network pruning perspective.
Here the structured pruning is adopted, which is based on the hypothesis that there is redundancy in the attention heads \cite{voita-etal-2019-Heavy,prasanna2020bertLottery}.
To obtain importance scores, we follow~\cite{michel2019sixteen,prasanna2020bertLottery} and calculate the expected sensitivity (gradient) of the attention heads to the mask variable $\xi^{(i, u)}$ (i.e., $\{0, 1\}$):
\vspace{-0.2cm}
\begin{equation}
\small
    \mathrm{MultiHeadAttn}(Q,K,V) = \mathrm{Concat}(\xi_1 head_1,.., \xi_{N_h} head_{N_h})
\end{equation}
% \vspace{-0.1cm}
\begin{equation}
\label{eq:cal_importance}
\small
S^{(i, u)}=E_{x \sim X}\left|\frac{\partial \mathcal{L}(x)}{\partial \xi^{(i, u)}}\right|
\end{equation}
where $X$ is the data distribution and $\mathcal{L}(x)$ the loss on sample $x$. Intuitively, if $S^{(i, u)}$ has a high value then changing $\xi^{(i, u)}$ is liable to have a large effect on the model (denotes the contribution score for attention head $i$ at layer $u$).
After calculating the contribution scores, we rank the attention heads of each Transformer encoder layer after layer normalization, and obtain those with the highest/lowest scores.

\begin{figure}[t!]
    \begin{center}
    \includegraphics[width=0.97\linewidth]{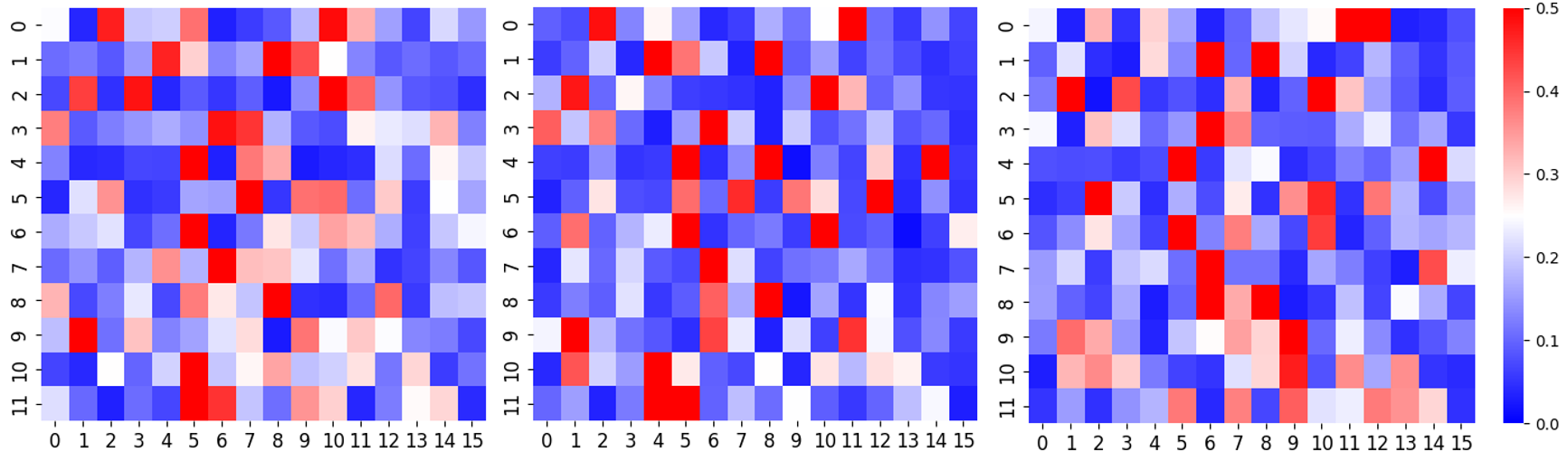}
    \end{center}
    \vspace{-0.6cm}
    \caption{Head importance heatmaps of the \textit{BART-large} model trained with three different training configurations.}
    \label{fig:headmap_compare}
\vspace{-0.3cm}
\end{figure}

\begin{figure}[t!]
    \begin{center}
    \includegraphics[width=0.75\linewidth]{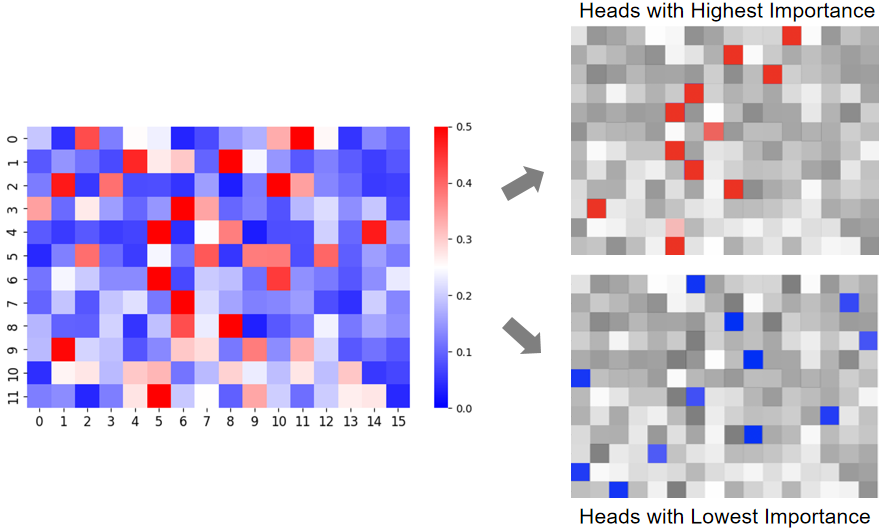}
    \end{center}
    \vspace{-0.6cm}
    \caption{Importance heatmap after averaging operation, and heads with highest/lowest scores are selected in each layer.}
    \label{fig:headmap_average}
\vspace{-0.3cm}
\end{figure}

In our preliminary fine-tuning experiments, we observed that there are some ranking variance. As shown in Figure \ref{fig:headmap_compare}, the head importance heatmaps of a \textit{BART-large} model \cite{lewis-etal-2020-bart} upon different training configurations of random seeds and learning rates are not exactly the same.
We then use averaging to exclude heads with high deviation. As shown in Figure \ref{fig:headmap_average}, some heads show consistently higher/lower layerwise importance scores; It indicates that the head importance has some correlation with the downstream task and training corpus.

To evaluate the impact of pruning heads on the downstream task at the inference stage, we prune the heads of a well-trained model on the summarization corpus, and compare results on the test set. As shown in Table \ref{table:pruning_result}, masking the highest-ranking heads of all Transformer layers leads to a 3.4\% relative decrease on ROUGE-L, while masking lowest-ranking heads only brings a 0.5\% drop.
To evaluate the impact of pruning heads at the training stage, we mask the attention heads based on their importance during the fine-tuning process. As shown in Table \ref{table:pruning_result}, the model can achieve a comparable result when we mask the highest-ranking heads of all layers. In contrast, the evaluation performance upon masking lowest-ranking (underused) heads even becomes slightly higher. We postulate that the model turns to exploit the rest heads.
Therefore, at both the training and inference stages, the head importance is effective to indicate its contribution to the task, and some heads provide limited contribution and can be pruned before fine-tuning.
\vspace{-0.2cm}

\begin{table}[t!]
\linespread{0.95}
\centering
\small
\resizebox{1.0\linewidth}{!}
{
\begin{tabular}{p{3.1cm}p{1.8cm}<{\centering}p{1.8cm}<{\centering}p{1.8cm}<{\centering}}
\toprule
      \textbf{Model}         & \textbf{ROUGE-1} & \textbf{ROUGE-2} & \textbf{ROUGE-L}  \\
\midrule
Baseline (\textit{BART-large})   &  53.14 &  28.58 &  49.69  \\
\midrule
\multicolumn{3}{l}{Pruning Heads at Inference Stage} \\
Highest-Ranking Heads  &  51.72 $[\downarrow $2.7\%$]$ &  27.10 $[\downarrow $5.2\%$]$ &  48.04 $[\downarrow $3.4\%$]$ \\
Lowest-Ranking Heads  &  52.89 $[\downarrow $0.5\%$]$ &  27.88 $[\downarrow $2.5\%$]$ &  49.46 $[\downarrow $0.5\%$]$ \\
\midrule
\multicolumn{3}{l}{Pruning Heads at Training Stage} \\
Highest-Ranking Heads  &  52.70 $[\downarrow $0.9\%$]$ &  28.05 $[\downarrow $1.9\%$]$ &  49.14 $[\downarrow $1.2\%$]$ \\
Lowest-Ranking Heads   &  53.16 $[\uparrow $0.1\%$]$ &  28.59 $[\uparrow $0.1\%$]$ &  49.73 $[\uparrow $0.1\%$]$ \\
\bottomrule
\end{tabular}
}
\caption{\label{table:pruning_result}ROUGE F1 scores on attention head pruning at training and inference stage. Relative changes are in brackets.}
\vspace{-0.4cm}
\end{table}

\section{Head Manipulation for Linguistic Feature Injection}
\subsection{Constructing Structure-Aware Matrix}
Given dialogue content after coreference resolution, to build the chain of a coreference cluster, we add links between each item and their mentions. Following previous works \cite{liu-etal-2021-coreference}, to better retain local information, we connect all adjacent items in one cluster. More specifically, given a cluster $C_i$ of $m$ items $\{E^i_{1},E^i_{2}...,E^i_{m}\}$, we add a bi-directional link of each $E$ to its precedent.
To construct a structure-aware matrix upon coreference chains for enhancing the Transformer encoder, here we investigate two methods:

\noindent\textbf{Full-link Matrix} Given a dialogue input $x$ of $n$ tokens (a sub-word tokenization is utilized), a structure-aware coreference matrix $A_x$ is initialized in a $n^2$ dimension. Iterating each coreference cluster $C$, the first token $t_i$ of each item (\emph{e.g.}, word and text span) is connected with the first token $t_j$ of its antecedent in the same cluster with a bi-directional edge (\emph{i.e.,} $A_{x}[i][j] = 1$ and $A_{x}[j][i] = 1$). Then the weights are re-scaled by averaging on the cluster size ($m$ items).

\noindent\textbf{Adjacent-link Matrix}
When the size of one cluster is big, its averaged weights in a full-link matrix will be very small and cause gradient vanishing. Therefore, following the feature aggregation from neighbors in graph neural networks (GNNs), we construct the adjacent-link matrix by only connecting each item with its nearest neighbors.

\begin{figure}[t!]
    \begin{center}
    \includegraphics[width=0.75\linewidth]{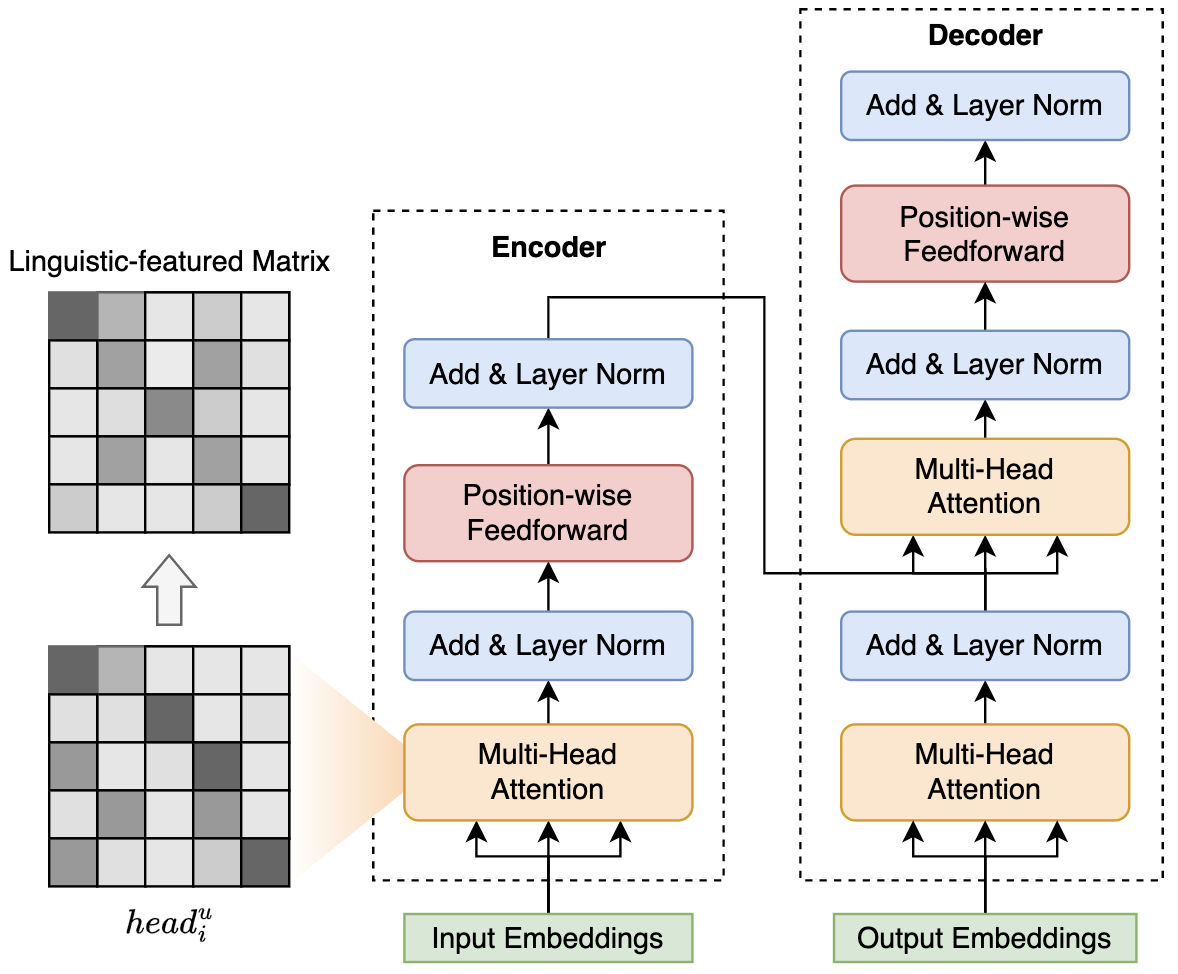}
    \end{center}
    \vspace{-0.5cm}
    \caption{Architecture overview of the coreference-aware Transformer with attention head manipulation.}
    \label{fig:transformer}
\vspace{-0.2cm}
\end{figure}

\begin{table}[t!]
\linespread{0.95}
\centering
\small
\resizebox{0.95\linewidth}{!}
{
\begin{tabular}{p{3.2cm}p{1.6cm}<{\centering}p{1.6cm}<{\centering}p{1.6cm}<{\centering}}
\toprule
      \textbf{Model}         & \textbf{ROUGE-1} & \textbf{ROUGE-2} & \textbf{ROUGE-L}  \\
\midrule
Baseline (\textit{BART-large})   &  53.14 &  28.58 &  49.69  \\
\textit{MV-BART-Large} &  53.42 &  27.98 &  49.97  \\
\textit{LM-Annotator ($\mathcal{D}_\mathrm{All}$)} &  53.70   &  28.79 &  50.81  \\
\midrule
\multicolumn{3}{l}{Probing-based Head Selection} \\
Full-link Matrix  &  53.80   &  28.58 &  50.25 \\
Adjacent-link Matrix   &   53.58   &   28.83  &  50.12 \\
\midrule
\multicolumn{3}{l}{Importance-based Head Selection} \\
Full-link Matrix  &  53.68   &  28.71 &  50.03  \\
Adjacent-link Matrix  &   53.98   &   29.15  &  50.73 \\
\bottomrule
\end{tabular}
}
\caption{\label{table:summ_result}F1 ROUGE scores on the abstractive dialogue summarization with attention head manipulation.}
\vspace{-0.3cm}
\end{table}

\subsection{Attention Head Manipulation}
After obtaining structure-aware matrices, we utilize them to enhance the Transformer-based summarizer. Here we directly manipulate attention heads with the featured weights, which is a parameter-free method and more computationally efficient than using additional neural components \cite{xu-etal-2020-discourse,raganato2020fixedPattern}. It saves 10M parameters (3.1\%) and 17\% inference time than a GNN-based model. As shown in Figure \ref{fig:transformer}, in $u$-th layer, if one head is lowest-ranking in the importance analysis, we modify it with weights from $A_x$ that present coreference information (\emph{i.e., Importance-based Head Selection}). In our setting, 6 of the 12 \textit{BART} encoding layers were processed for hierarchical modeling, and we empirically found that manipulating heads in higher layers performed better.
Additionally, we compare the proposed method to another one named \emph{Probing-based Head Selection} \cite{liu-etal-2021-coreference}, where attention heads that are most similar to the structure-aware matrix $A_x$ (using cosine similarity as measurement) are selected for feature injection (There are no overlapped heads of these two selection strategies).

\subsection{Results on Dialogue Summarization}
Aside from the base model \textit{BART-large} \cite{lewis-etal-2020-bart}, we include two recent state-of-the-art models: \textit{MV-BART-Large} \cite{chen-yang-2020-multi} and \textit{LM-Annotator ($D_{All}$)} \cite{feng2021survey} for extensive comparison.
As shown in Table \ref{table:summ_result}, incorporating coreference information helps the backbone \textit{BART-large}, and makes it comparable to the state-of-the-art models that use additional training data and neural components. In particular, importance-based head selection with adjacent-link matrix performed best with 1.6\%, 2.0\%, and 2.1\% relative F1 score improvement, which is better than the full-link scheme, and the probing-based approach.

\subsection{Importance Analysis of Manipulated Heads}
\begin{figure}[h]
    \begin{center}
    \includegraphics[width=0.85\linewidth]{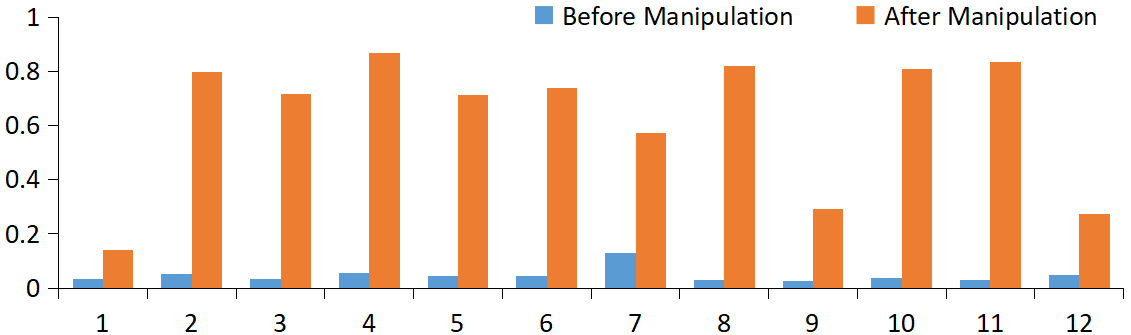}
    \end{center}
    \vspace{-0.5cm}
    \caption{Head importance comparison of before and after the feature injection. Axis X denotes the layer number. Axis Y denotes the normalized importance score.}
    \label{fig:before_after}
\vspace{-0.4cm}
\end{figure}

To qualitatively assess the effectiveness of head manipulation, we conduct an ablation study via head pruning. At the inference stage, we mask the heads that are injected with structure-aware coreference features, and compare it with the unaltered model. As shown in Table \ref{table:ablation_result}, pruning the manipulated heads leads to significant performance drop, and the model is affected more by the importance-based than probing-based strategy.
Moreover, we compare the importance scores (see Eq. \ref{eq:cal_importance}) of before and after head manipulation in all encoder layers, and observe that the previously underused heads weigh much higher, demonstrating that the enhanced model effectively utilize the injected features (see Figure \ref{fig:before_after}).
\vspace{-0.1cm}

\begin{table}[t!]
\linespread{0.95}
\centering
\small
\resizebox{1.0\linewidth}{!}
{
\begin{tabular}{p{3.0cm}p{1.8cm}<{\centering}p{1.8cm}<{\centering}p{1.8cm}<{\centering}}
\toprule
      \textbf{Model}         & \textbf{ROUGE-1} & \textbf{ROUGE-2} & \textbf{ROUGE-L}  \\
\midrule
\multicolumn{3}{l}{Inference Pruning of Probing-based Heads} \\
Full-link Matrix  &  52.92 $[\downarrow $1.7\%$]$  &  28.01 $[\downarrow $2.0\%$]$ &  49.17 $[\downarrow $2.2\%$]$ \\
Adjacent-link Matrix  &   52.75 $[\downarrow $1.6\%$]$  &   28.10 $[\downarrow $2.5\%$]$ &  49.35 $[\downarrow $1.6\%$]$ \\
\midrule
\multicolumn{3}{l}{Inference Pruning of Importance-based Heads} \\
Full-link Matrix  &  52.32 $[\downarrow $2.5\%$]$  &  27.41 $[\downarrow $4.4\%$]$ &  48.47 $[\downarrow $3.1\%$]$ \\
Adjacent-link Matrix  &   52.44 $[\downarrow $2.9\%$]$  &   27.75 $[\downarrow $4.9\%$]$ &  48.60 $[\downarrow $4.2\%$]$ \\
\bottomrule
\end{tabular}
}
\caption{\label{table:ablation_result}Ablation study via head pruning at inference stage.}
\vspace{-0.4cm}
\end{table}

\section{Conclusion}
In this work, we revisited the attention head selection strategy for feature injection from a network pruning perspective. Head importance scoring and ranking of a Transformer-based summarizer showed that some heads are underused after task-specific training. We then manipulated such heads to incorporate structure-aware dialogue coreference features.
Experimental results showed that the importance-based head selection is effective for linguistic knowledge injection, and incorporating coreference information is beneficial for dialogue summarization. As a general and computationally efficient approach, this can also be extended to other Transformer-based models and natural language tasks.

\vfill\pagebreak

% \section{REFERENCES}
% \label{sec:refs}

% List and number all bibliographical references at the end of the
% paper. The references can be numbered in alphabetic order or in
% order of appearance in the document. When referring to them in
% the text, type the corresponding reference number in square
% brackets as shown at the end of this sentence \cite{C2}. An
% additional final page (the fifth page, in most cases) is
% allowed, but must contain only references to the prior
% literature.

% References should be produced using the bibtex program from suitable
% BiBTeX files (here: strings, refs, manuals). The IEEEbib.bst bibliography
% style file from IEEE produces unsorted bibliography list.
% -------------------------------------------------------------------------
\bibliographystyle{IEEEbib}
\bibliography{refs}

\begin{thebibliography}{10}

\bibitem{dasigi-etal-2019-quoref}
Pradeep Dasigi, Nelson~F. Liu, Ana Marasovi{\'c}, Noah~A. Smith, and Matt Gardner,
\newblock ``{Q}uoref: A reading comprehension dataset with questions requiring coreferential reasoning,''
\newblock in {\em Proceedings of the EMNLP-IJCNLP 2019}, Hong Kong, China, Nov. 2019, pp. 5925--5932, Association for Computational Linguistics.

\bibitem{vaswani-2017-Transformer}
Ashish Vaswani, Noam Shazeer, Niki Parmar, Jakob Uszkoreit, Llion Jones, Aidan~N Gomez, {\L}ukasz Kaiser, and Illia Polosukhin,
\newblock ``Attention is all you need,''
\newblock in {\em Proceedings of the NeurIPS}, 2017.

\bibitem{tenney2018what}
Ian Tenney, Patrick Xia, Berlin Chen, Alex Wang, Adam Poliak, R~Thomas McCoy, Najoung Kim, Benjamin~Van Durme, Sam Bowman, Dipanjan Das, and Ellie Pavlick,
\newblock ``What do you learn from context? probing for sentence structure in contextualized word representations,''
\newblock in {\em Proceedings of the ICLR 2019}, 2019.

\bibitem{hewitt-manning-2019-structural}
John Hewitt and Christopher~D. Manning,
\newblock ``{A} structural probe for finding syntax in word representations,''
\newblock in {\em Proceedings of the NAACL 2019}, Minneapolis, Minnesota, June 2019, pp. 4129--4138.

\bibitem{xu-etal-2020-discourse}
Jiacheng Xu, Zhe Gan, Yu~Cheng, and Jingjing Liu,
\newblock ``Discourse-aware neural extractive text summarization,''
\newblock in {\em Proceedings of the ACL 2020}, Online, July 2020, pp. 5021--5031, Association for Computational Linguistics.

\bibitem{liu-etal-2021-coreference}
Zhengyuan Liu, Ke~Shi, and Nancy Chen,
\newblock ``Coreference-aware dialogue summarization,''
\newblock in {\em Proceedings of the SIGDIAL 2021}, Singapore and Online, July 2021, pp. 509--519, Association for Computational Linguistics.

\bibitem{wang-etal-2019-treeTransformer}
Yaushian Wang, Hung-Yi Lee, and Yun-Nung Chen,
\newblock ``Tree transformer: Integrating tree structures into self-attention,''
\newblock in {\em Proceedings of the EMNLP-IJCNLP 2019}, Hong Kong, China, Nov. 2019, pp. 1061--1070.

\bibitem{voita-etal-2019-Heavy}
Elena Voita, David Talbot, Fedor Moiseev, Rico Sennrich, and Ivan Titov,
\newblock ``Analyzing multi-head self-attention: Specialized heads do the heavy lifting, the rest can be pruned,''
\newblock in {\em Proceedings of the ACL 2019}, Florence, Italy, July 2019, pp. 5797--5808.

\bibitem{michel2019sixteen}
Paul Michel, Omer Levy, and Graham Neubig,
\newblock ``Are sixteen heads really better than one?,''
\newblock {\em Advances in Neural Information Processing Systems}, vol. 32, 2019.

\bibitem{gliwa-etal-2019-samsum}
Bogdan Gliwa, Iwona Mochol, Maciej Biesek, and Aleksander Wawer,
\newblock ``{SAMS}um corpus: A human-annotated dialogue dataset for abstractive summarization,''
\newblock in {\em Proceedings of the 2nd Workshop on New Frontiers in Summarization}, Hong Kong, China, Nov. 2019, pp. 70--79, Association for Computational Linguistics.

\bibitem{lewis-etal-2020-bart}
Mike Lewis, Yinhan Liu, Naman Goyal, Marjan Ghazvininejad, Abdelrahman Mohamed, Omer Levy, Veselin Stoyanov, and Luke Zettlemoyer,
\newblock ``{BART}: Denoising sequence-to-sequence pre-training for natural language generation, translation, and comprehension,''
\newblock in {\em Proceedings of the ACL 2020}, Online, July 2020, pp. 7871--7880, Association for Computational Linguistics.

\bibitem{liu2019topic}
Zhengyuan Liu, Angela Ng, Sheldon Lee, Ai~Ti Aw, and Nancy~F Chen,
\newblock ``Topic-aware pointer-generator networks for summarizing spoken conversations,''
\newblock in {\em 2019 IEEE Automatic Speech Recognition and Understanding Workshop (ASRU)}. IEEE, 2019, pp. 814--821.

\bibitem{feng2021survey}
Xiachong Feng, Xiaocheng Feng, and Bing Qin,
\newblock ``A survey on dialogue summarization: Recent advances and new frontiers,''
\newblock {\em arXiv preprint arXiv:2107.03175}, 2021.

\bibitem{SACKS19787}
HARVEY Sacks, EMANUEL~A. SCHEGLOFF, and GAIL JEFFERSON,
\newblock ``A simplest systematics for the organization of turn taking for conversation,''
\newblock in {\em Studies in the Organization of Conversational Interaction}, JIM SCHENKEIN, Ed., pp. 7 -- 55. Academic Press, 1978.

\bibitem{jurafsky2008speech}
Daniel Jurafsky and James~H Martin,
\newblock ``Speech and language processing: An introduction to speech recognition, computational linguistics and natural language processing,''
\newblock {\em Upper Saddle River, NJ: Prentice Hall}, 2008.

\bibitem{tang-etal-2022-confit}
Xiangru Tang, Arjun Nair, Borui Wang, Bingyao Wang, Jai Desai, Aaron Wade, Haoran Li, Asli Celikyilmaz, Yashar Mehdad, and Dragomir Radev,
\newblock ``{CONFIT}: Toward faithful dialogue summarization with linguistically-informed contrastive fine-tuning,''
\newblock in {\em Proceedings of the NAACL 2022}, Seattle, United States, July 2022, pp. 5657--5668.

\bibitem{lin-och-2004-automatic}
Chin-Yew Lin and Franz~Josef Och,
\newblock ``Automatic evaluation of machine translation quality using longest common subsequence and skip-bigram statistics,''
\newblock in {\em Proceedings of the ACL 2004}, Barcelona, Spain, July 2004.

\bibitem{prasanna2020bertLottery}
Sai Prasanna, Anna Rogers, and Anna Rumshisky,
\newblock ``When bert plays the lottery, all tickets are winning,''
\newblock in {\em Proceedings of the EMNLP 2020}, 2020, pp. 3208--3229.

\bibitem{raganato2020fixedPattern}
Alessandro Raganato, Yves Scherrer, and J{\"o}rg Tiedemann,
\newblock ``Fixed encoder self-attention patterns in transformer based machine translation,''
\newblock in {\em Findings of the EMNLP 2020}. 2020, pp. 556--568, Association for Computational Linguistics.

\bibitem{chen-yang-2020-multi}
Jiaao Chen and Diyi Yang,
\newblock ``Multi-view sequence-to-sequence models with conversational structure for abstractive dialogue summarization,''
\newblock in {\em Proceedings of the EMNLP 2020}, Online, Nov. 2020, pp. 4106--4118, Association for Computational Linguistics.

\end{thebibliography}

\end{document}